\documentclass[letterpaper, 10 pt, conference]{ieeeconf}  

\IEEEoverridecommandlockouts                              
 
\usepackage{graphics} 
\usepackage{epsfig} 
\usepackage{times} 
\usepackage{amsmath} 
\usepackage{amssymb}  
\usepackage[utf8]{inputenc} 
\usepackage[T1]{fontenc}    
\usepackage{hyperref}       
\usepackage{url}            
\usepackage{booktabs}       
\usepackage{amsfonts}       
\usepackage{nicefrac}       
\usepackage{microtype}      
\usepackage{cleveref}
\usepackage{graphicx}
\usepackage{float}
\usepackage{multirow}
\usepackage{booktabs}
\usepackage{booktabs}
\usepackage{multirow}
\usepackage[table]{xcolor}
\definecolor{lightgray}{gray}{0.9}
\usepackage{diagbox}
\usepackage{algorithm}
\usepackage{algpseudocode}
\usepackage{arydshln}

\usepackage{listings}

\title{\LARGE \bf Boosting Universal LLM Reward Design through Heuristic Reward Observation Space Evolution}
\author{Zen Kit Heng$^{\dag}$, Zimeng Zhao$^{\dag}$, Tianhao Wu$^{\ddag}$, Yuanfei Wang$^{\ddag}$, Mingdong Wu, Yangang Wang, Hao Dong
\thanks{Zen Kit Heng, Tianhao Wu, Yuanfei Wang, Mingdong Wu and Hao Dong are with the Center on Frontiers of Computing Studies, School
of Computer Science, Peking University, Beijing 100871, China, also with
PKU-Agibot Lab, School of Computer Science, Peking University, Beijing
100871, China, and also with National Key Laboratory for Multimedia
Information Processing, School of Computer Science, Peking University,
Beijing 100871, China. Zimeng Zhao and Yangang Wang are with School of Automation, Southeast University, Nanjing, China.}
\thanks{$^{\dag}$ The first two authors contributed equally. }
\thanks{$^{\ddag}$ The second two authors contributed equally. }
\thanks{Corresponding to hao.dong@pku.edu.cn}
}

\def\wrt{{w}.{r}.{t}.}

\def\forexample{\emph{e.g}.}

\def\figmk{Fig.~}

\def\equationmk{Eqn.~}
\def\supmat{\href{https://jingjjjjjie.github.io/LLM2Rewards}{website}}

\def\rwdobs{\textbf{ROS}}

\begin{document}

\maketitle
\thispagestyle{empty}
\pagestyle{empty}

\begin{abstract}
    Large Language Models (LLMs) are emerging as promising tools for automated reinforcement learning (RL) reward design, owing to their robust capabilities in commonsense reasoning and code generation. By engaging in dialogues with RL agents, LLMs construct a Reward Observation Space (ROS) by selecting relevant environment states and defining their internal operations. However, existing frameworks have not effectively leveraged historical exploration data or manual task descriptions to iteratively evolve this space. In this paper, we propose a novel heuristic framework that enhances LLM-driven reward design by evolving the ROS through a table-based exploration caching mechanism and a text-code reconciliation strategy. Our framework introduces a state execution table, which tracks the historical usage and success rates of environment states, overcoming the Markovian constraint typically found in LLM dialogues and facilitating more effective exploration. Furthermore, we reconcile user-provided task descriptions with expert-defined success criteria using structured prompts, ensuring alignment in reward design objectives. Comprehensive evaluations on benchmark RL tasks demonstrate the effectiveness and stability of the proposed framework. Code and video demos are available at \url{https://jingjjjjjie.github.io/LLM2Rewards}
\end{abstract}


\section{Introduction} 
Simulation is of paramount importance in scaling up robotic learning, as it provides more controllable environments for acquiring diverse robotic skills \cite{singh2009rewards} and offers additional guidance for effective sim-to-real transfer~\cite{ma2024dreureka}. Traditional methods rely on manual reward shaping, which is labor-intensive and may not generalize well across different tasks and environments~\cite{ng1999policy}. Moreover, poorly designed rewards can hinder the learning process or lead to unintended behaviors, complicating the development of robust robotic systems~\cite{leike2018scalable}.

Recent advancements in Large Language Models (LLMs)~\cite{achiam2023gpt} have opened new avenues for automated reward design. LLMs are believed to encapsulate vast amounts of human knowledge and logic from their training corpus. Consequently, the embodied AI community regards LLMs as generalizable encyclopedias, guiding agents to learn skills universally across diverse environment-robot settings under an imitation learning~\cite{ha2023scaling,belkhale2024rt} or reinforcement learning~\cite{yu2023language,ma2023eureka,wang2023robogen,xiao2023unified} formulation.

Focusing on the latter, this work aims to build a universal RL reward design scheme with LLMs. The primary challenge lies in grounding~\cite{ahn2022can}: making RL agents' perceptions understandable for LLMs' planning and ensuring that LLMs' code is actionable for RL agents' control. Formally, LLMs design an RL task reward through constructing a corresponding \emph{Reward Observation Space} (\rwdobs). This space contains a subset of all available environment-robot states, along with the operations defined upon those subset members. Our key idea is to regard the LLM reward design problem as a heuristic sampling to evolve the \rwdobs~ driven by both curiosity and success feedback.

\begin{figure}[t]\label{Teaser}
    \begin{center}
    \includegraphics[width=\linewidth]{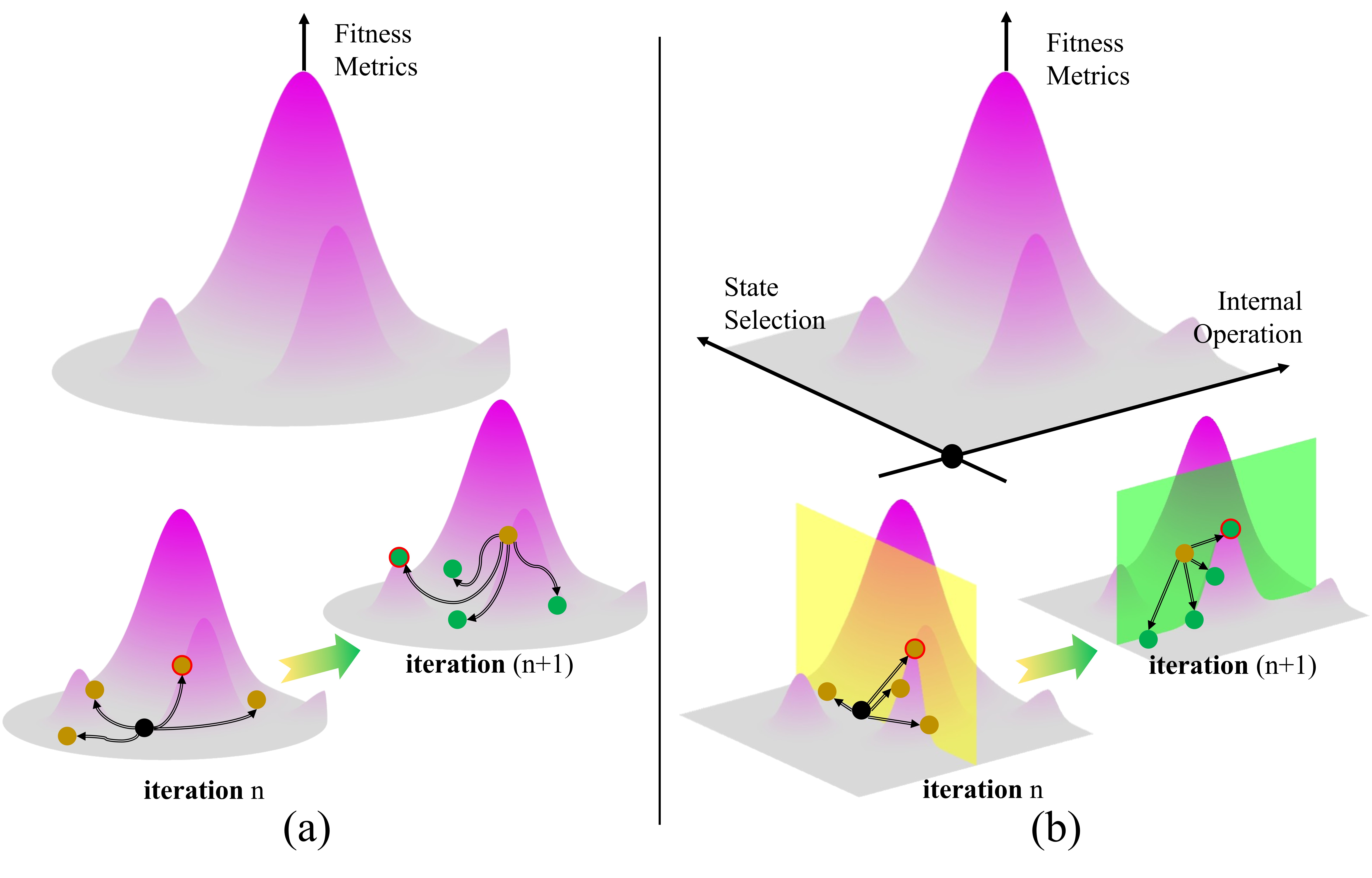}
    \end{center}
    \vspace{-0.3cm}
    \caption{\textbf{Comparison diagram of evolutionary process.}(a) Eureka's evaluation and sampling. (b) Our evaluation and sampling.}
    \vspace{-0.7cm}
    \label{fig:teasor}
\end{figure}

Limited by LLM context length, pioneer works~\cite{ma2023eureka, zeng2024learning} only adopted the local optima \rwdobs~ of the last iteration as a sampling guidance in the current iteration. In pursuit of more thorough and efficient communications, we first incorporate a state execution table accessible to all iterations and samples, breaking the Markovian constraint in LLM dialogues. This table keeps track of historical usage frequency and success contributions of each state in the RL task, ensuring that novel \rwdobs~configurations are explored by encouraging LLMs to prioritize states with higher success contributions and fewer usages. To further optimize this process, we employ the following strategies: (i) We disentangle the design of space member selection and internal member operation into two sub-problems, alternating between them across iterations, reducing complexity and improving comparability within \rwdobs~during each iteration. (ii) To prevent LLMs from being misled by local optima, the full reward code is only provided if the success rate exceeds an adaptive threshold. Otherwise, the reward code is truncated~\cite{omer2023llms}. (iii) Execution errors are mitigated by only updating the state execution table with successful runs, keeping LLMs focused on valid state configurations.

We further get insights on conveying structured and non-controversial information~\cite{xiao2023unified,yu2023language,yang2023set} to LLMs. Current formulation~\cite{ma2023eureka, zeng2024learning} takes the design mission from a \emph{user's} description text, while iteratively evaluating each design sample according to another \emph{expert's} success code. As a potential risk, the intentions of these two individuals may be misaligned or even contradictory. To this end, we resort to LLM to reconcile their potential contradictions by prompting with their structured templates. Surprisingly, even for a new task without a definition of success, our framework could facilitate LLM in writing an expert-level success function according to the above templates and user task description. Afterward, the reconciled description becomes the LLM's mission, and the reconciled success replaces the vanilla success. For a fair comparison with the existing approaches~\cite{ma2023eureka,yu2023language,zeng2024learning}, our LLM preprocessing is isolated from the subsequent design iterations and the reconciled success is still invisible during LLM design. In our comparison, their performance on reconciled settings has also been evaluated. 

In summary, our main contributions are:

\noindent$\bullet$ A heuristic \rwdobs ~evolution framework, boosting the performance of LLM reward design. 

\noindent$\bullet$ A table-based exploration caching mechanism, breaking the Mar0kovian constraint in LLM dialogues. 

\noindent$\bullet$ A user-expert reconciliation strategy, filling the cognitive gap between users and experts to the same task.
\section{Related work}

\noindent\textbf {Reward Design Problem.}
Rewards are the primary mechanism through which agents learn desirable behaviors in a given environment~\cite{dayan2002reward}, and their formulation critically affects the efficiency and success of the learning process~\cite{kim2021observation}. Historically, reward functions have been hand-crafted by domain experts for specific tasks like robot navigation or games~\cite{mnih2015human,silver2016mastering}. However, crafting effective reward functions is challenging, as it requires balancing exploration with enough guidance to avoid suboptimal behavior~\cite{sutton2018reinforcement}. For more complex tasks, such as robotic manipulation involving continuous control, reward design becomes significantly more challenging~\cite{akkaya2019solving,chen2022towards}. To address the above challenges, researchers have explored automatic reward generation techniques. Inverse reinforcement learning (IRL)~\cite{arora2021survey,10.5555/645529.657801} aims to infer reward functions from expert behavior rather than relying on manual design. IRL has demonstrated success across various domains, yet it often demands high-quality expert demonstrations, which can be costly and time-consuming to obtain. Moreover, the inferred reward functions are not always unique or interpretable, leading to ambiguity when applied to new contexts~\cite{ziebart2008maximum}. Evolutionary algorithms represent another approach to reward design. Early works, such as \cite{niekum2010genetic}, utilized these algorithms to iteratively optimize reward functions by selecting the most promising candidates. Although promising, these methods are computationally intensive and require predefined templates for possible reward structures, which constrains their flexibility. The recent advent of large language models (LLMs)~\cite{openai2023gpt4,white2023chatgpt} has expanded the potential for automatic reward generation. For example, Eureka framework~\cite{ma2023eureka} generates reward functions from raw environment descriptions and task code. However, the quality of rewards produced by Eureka can be inconsistent, and the generation process may suffer from instability particularly when applied to tasks requiring fine-tuned control. By incorporating advanced evolutionary algorithms, our approach ensures stable reward generation, especially in environments with a high degree of freedom action space~\cite{chen2022towards}.

\noindent\textbf {Scaling up Robotic Learning in Simulations.} 
Simulations provide a controlled environment where robots can acquire diverse skills with minimal risk and cost, making it a fundamental tool for research in embodied AI~\cite{tobin2017domain}. However, despite the advances in simulation platforms, scaling up robotic learning presents several hurdles, primarily in the generation of diverse tasks. Automating task generation is therefore a vital step toward scaling up learning processes in simulation. A key issue in scaling up robotic learning is the creation of diverse and meaningful tasks within simulation environments. Traditionally, task generation has relied heavily on manual design, where experts must define each task, its associated assets, and the reward functions that guide the robot’s learning process. This approach is not only labor-intensive but also limits the scope and diversity of tasks that can be generated. One promising direction to scale up robotic learning is through multi-task and meta-reinforcement learning~\cite{james2020rlbench,yu2020meta,chen2022towards}. In multi-task RL, robots learn multiple tasks simultaneously, sharing knowledge across tasks to improve generalization. Frameworks like Meta-World \cite{yu2020meta} provides a collection of manipulation tasks that encourage robots to develop transferable skills. Meta-reinforcement learning goes further by training robots to learn how to learn, enabling them to quickly adapt to new tasks based on previous experience. Some attempts~\cite{finn2017model} aims to optimize a robot’s policy in such a way that it can rapidly adapt to new, unseen tasks with minimal additional training. These approaches are essential for developing generalist robots that can operate in diverse, dynamic environments. ~~This work aims to create diverse, user-described tasks within existing simulation environments, promoting multi-task learning and expanding the applicability of current simulations.
\begin{figure*}[t]
    \begin{center}
        \includegraphics[width=0.8\linewidth]{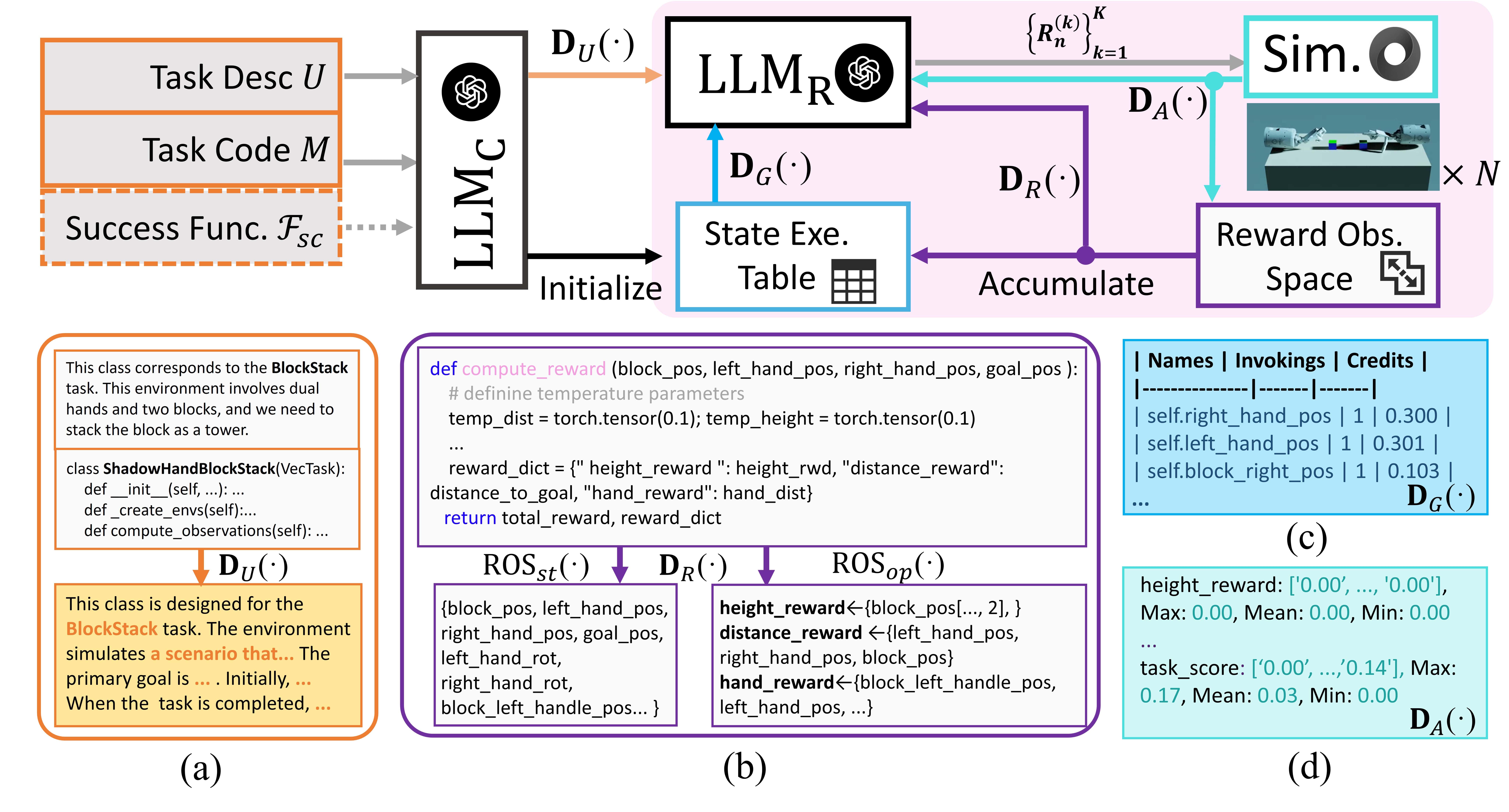}
    \end{center}
    \vspace{-8mm}
    \caption{The pipeline of our proposed framework for heuristic Reward Observation Space (\rwdobs) evolution in LLM-aided RL reward design. (a) User-expert Mission Reconciling. (b) Observation Space Disentanglement. (c) Reward State Execution. (d) Reward Item Performance. }
    \vspace{-0.5cm}
    \label{fig:pipeline}
\end{figure*}

\section{Problem Formulation}

We address the reward design problem (RDP) for general RL environments~\cite{singh2009rewards}, which can be formalized as a tuple $\mathrm{RDP} \triangleq \langle M, {R}, \varLambda, \mathcal{F}_{ft}\rangle$. Given a world model $M$, the optimal reward function ${R}^\star$ is expected to efficiently guide an RL algorithm $\varLambda$ to accomplish the task:
\begin{equation}
    \begin{aligned}
        {R}^\star = \arg\max_{{R}}\mathcal{F}_{ft}[\varLambda({R})]
    \end{aligned}
    \label{eqn_rdp}
\end{equation}

\noindent\textbf{World model} $M$ defines an environment-robot setting including state $S\in\mathcal{S}$, action $A\in\mathcal{A}$ and transition $\mathcal{S}\times\mathcal{A}\mapsto\mathcal{S}$. A task $\varGamma_{M}$ defined in $M$ is to achieve a clear goal. From user's perspective, this goal is expressed through the task description $U$(\forexample ``make humanoids stand up''). From expert's perspective, it is grounded as the success function $\mathcal{F}_{sc}:\mathcal{S}\mapsto\{0,1\}$ (\forexample ``take 1 when the humanoid torso is above a certain height otherwise 0''). 

\noindent\textbf{Reward function} ${R}:\mathcal{S}\times\mathcal{A}\mapsto\mathbb{R}$ compactly quantifies the contribution of every $\langle S,A \rangle$ pattern to $\varGamma_{M}$ with a scalar. When ${R}$ is represented by a logic code, its input is a subset of $\mathcal{S}$, and its output scalar is obtained by operations within the subset states. 

\noindent\textbf{RL algorithm} $\varLambda$ explores the optimal policy $\pi=\pi_\varLambda^\star$ guided by a form of ${R}$ to complete $\varGamma_{M}$. An interaction episode between $\pi_\varLambda$ and the $M$' transition results in a Markov Decision Process. The huge exploration space causes $\varLambda$ to have no guarantee that $\pi_\varLambda^\star$ will be found on every run, even if the same ${R}$ is adopted.

\noindent\textbf{Fitness function} $\mathcal{F}_{ft}: \varLambda({R})\mapsto\mathbb{R}$ crucially controls the optimization in \equationmk\ref{eqn_rdp} by measuring how well a form of ${R}$ can guide $\varLambda$ to complete $\varGamma_{M}$. A na\"ive setting is $\mathcal{F}_{ft}=\mathcal{F}_{sc}$. However, the optimization guided solely by $\mathcal{F}_{sc}$ lack effectiveness since $\mathcal{F}_{sc}$ is usually sparse~\cite{ma2023eureka}. 

In essence, RDP~\equationmk\ref{eqn_rdp} requires an expert-crafted $\mathcal{F}_{ft}$ to guide the optimization of $R$ effectively.

\section{Methodology}
\vspace{-1mm}
In this paper, we treat the dialogue between LLM and $\varLambda({R})$ as an iterative evolutionary process similar to \equationmk\ref{eqn_rdp}. This allows users without domain expertise to design rewards using natural language $U$ while extending the guidance beyond $\mathcal{F}_{ft}$ alone. Specifically, the LLM is prompted to design $K$ reward samples $\mathcal{R}_n \triangleq \{R^{(k)}_n\}_{k=1}^K$ at the $n$-th iteration ($n=1,...,N$):
\begin{equation}
    \begin{aligned}
        \mathcal{R}_n &= \text{LLM}_{R} \left( \mathbf{D}_R(R_{n-1}^\star), \mathbf{D}_A[\varLambda({R_{n-1}^\star})], \right. \\
        &\quad \left. \mathbf{D}_G(\sum^{n-1}_{m}\mathcal{R}_m)~|~ \mathbf{D}_U(U) \right) \\
        R_{n}^\star &= \arg\max_{R \in \mathcal{R}_n}\mathcal{F}_{sc}[\varLambda({R})] \\
    \end{aligned}
    \label{eqn_llm}
\end{equation}
Initially, $\mathbf{D}_U(\cdot)$ interprets the task description $U$ (\ref{subsec_DU}). In subsequent iterations (\ref{subsec_DADRDG}), composite guidance is applied. $\mathbf{D}_R(\cdot)$ maps ${R}$ into its \emph{Reward Observation Space} (\rwdobs). $\mathbf{D}_A(\cdot)$ summarizes the performance of $\varLambda$. $\mathbf{D}_G(\cdot)$ acts as memory, transcribing all historical reward samples into a \emph{State Execution Table}. The target $R^\star$ is approximated by selecting the best-performing local optimum samples over $N$ iterations:
\begin{equation}
    \begin{aligned}
        R^\star &\approx \arg\max_{R \in \{R_{1}^\star, ..., R_{N}^\star\}}\mathcal{F}_{sc}[\varLambda({R})]
    \end{aligned}
    \label{eqn_llm2}
\end{equation}

\subsection{User-expert Mission Reconciling}
\label{subsec_DU}
Two types of human involvement influence \ref{eqn_llm}: users describe tasks through $U$, while experts define success criteria via $\mathcal{F}_{sc}$. Conflicts between $U$ and $\mathcal{F}_{sc}$ can confuse $\text{LLM}_{R}$. To mitigate this, we employ another LLM ($\text{LLM}_{C}$) to reconcile these potential discrepancies. Importantly, $\text{LLM}_{C}$ does not share context with $\text{LLM}_{R}$, ensuring that the code of $\mathcal{F}_{sc}$ is invisible to $\text{LLM}_{R}$~\cite{ma2023eureka}.

\noindent\textbf{User description structuring.} Compared to $\mathcal{F}_{sc}$, $U$ can be more uncertain and flexible. To standardize $U$, we introduce a task-independent template $T_U$:
\begin{equation}
    \begin{aligned}
        \mathbf{D}_U(U) \triangleq \text{LLM}_{C} (U, \mathcal{F}_{sc}, M ~|~ T_U)
    \end{aligned}
    \label{eqn_DU}
\end{equation}
After this prompting, $\mathbf{D}_U(U)$ is expanded to include: (i) the composition of robots and objects in $M$. (ii) the goal states for these objects. (iii) the initial conditions, and (iv) potential post-goal states (see \supmat ~for details on $T_U$).

\noindent\textbf{Expert knowledge transfer.} When no success definition exists for a task in $M$, $\text{LLM}_{C}$ is further prompted to write the code for $\mathcal{F}_{sc}$:
\begin{equation}
    \begin{aligned}
        \left[\mathbf{D}_U(U), \mathcal{F}_{sc}\right] \triangleq \text{LLM}_{C} (U, M ~|~ T_U, \mathbf{D}_U(U^\prime), \mathcal{F}_{sc}^\prime)
    \end{aligned}
    \label{eqn_DU2}
\end{equation}
where $(U^\prime, \mathcal{F}_{sc}^\prime)$ corresponds to another task defined on $M$. This paradigm increases the reusability of $M$, and also makes our framework less dependent on the existence of $\mathcal{F}_{sc}$. 

\subsection{Observation Space Evaluation}
\label{subsec_DADRDG}

In subsequent iterations, $\text{LLM}_{R}$ is mainly guided by the following 3 types of guidance: (i) $\mathbf{D}_R(R_{n-1}^\star)$ acts as the reward example. (ii) $\mathbf{D}_A[\varLambda({R_{n-1}^\star})]$ reflects the detailed training effects of this example. (iii) $\mathbf{D}_G(\sum^{n-1}_{m}\mathcal{R}_m)$ keeps the historical exploration memory of $\text{LLM}_{R}$. 

\noindent\textbf{Reward Observation Space.}We argue that the exploration ability of $\text{LLM}_{R}$ iterations would be compromised when a code example from the last iteration is fed directly into the context of the current iteration. Especially when $R_{n-1}^\star$ is mediocre, it would cause $\text{LLM}_{R}$ to repeat this situation. To address this problem, we introduce the concept of \emph{Reward Observation Space} (\rwdobs). This space contains a subset of all available environment-robot states $\text{ROS}_{st}(R)$, along with the operations defined upon those subset members $\text{ROS}_{op}(R)$. This makes the information in a reward more compact and structured than line-by-line code. 

\begin{figure}[!t]
    \centering
    \includegraphics[width=\linewidth]{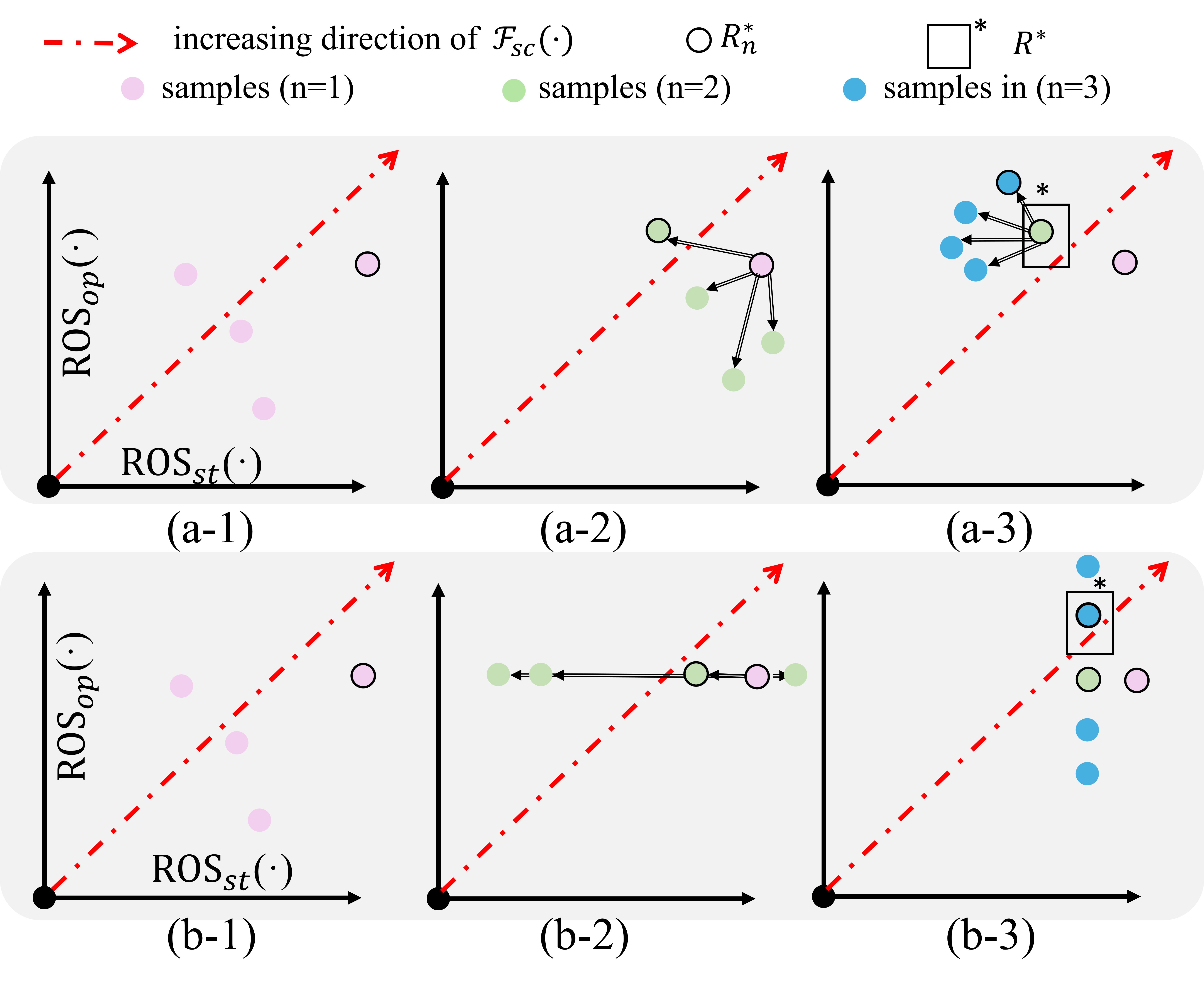}
    \vspace{-8mm}
    \caption{\textbf{Schematic illustration of the difference in sampling process for different on reward space$\mathcal{R}$.} Compared to Eureka~\cite{ma2023eureka}, observation Space disentanglement improves the efficiency of the sampling process of LLM by reducing the degrees of freedom.}
    \vspace{-4mm}
    \label{fig_oddeven}
\end{figure}

\noindent\textbf{Observation Space disentanglement.} 
As shown in \figmk\ref{fig_oddeven}, we further disentangle the design of the \rwdobs~ into two sub-problems: space member selection $\text{ROS}_{st}(R)$ and internal member operation $\text{ROS}_{op}(R)$. Consequently, the way $\mathbf{D}_R(\cdot)$ acts on $R^\star_n$ varies in different situations: 
\begin{equation}
    \mathbf{D}_R(R^\star_n) = \left\{
    \begin{aligned}
        \text{ROS}_{st}&(R^\star_n), \text{for odd}~n~\text{or}\\
        &~ \mathcal{F}_{sc}[\varLambda({R_{n}^\star})]< \tau \\
        \text{ROS}_{op}&(R^\star_n), \text{otherwise}~ \\
    \end{aligned}
    \right.
    \label{eqn_DR}
\end{equation}
The space member selection is triggered when the iteration index $n$ is odd or the success score is less than a task-specific threshold $\tau$. In this situation, $LLM_R$ is encouraged to select states from $M$ that differ from $\text{ROS}_{st}(R^\star_n)$ as the observation members. On the other hand, the optimization to internal member operation is triggered when the success achieving by $R^\star_n$ is considerable within even iterations. $LLM_R$ is prompted to use the same observation members in $\text{ROS}_{st}(R^\star_n)$ to devise novel reward items. This disentanglement strategy not only improves the efficiency of the reward design process but also enhances the comparability of different \rwdobs~ configurations within a single iteration, leading to more consistent and reliable reward outcomes. 

\noindent\textbf{State execution table.} 
Considering the contradiction between the Markovian nature and the token consumption constrains of an LLM dialogue, most approaches~\cite{ma2023eureka,ahn2022can} keep LLM memory by resending part of the history as additional input. By contrast, we summarize the full history in a table form $\text{SET}$ with less and constant token consumption. The column definition of $\text{SET}$ is shown in panel (c) of \figmk\ref{fig:pipeline}: Column 1 enumerates the names of all states in $M$. Column 2 indicates how many times each state is adopted in all historical rewards. Column 3 indicates the contribution to the task success of each state. This is calculated by dividing $\mathcal{F}_{sc}[\varLambda(R)]$ evenly to each state contained in $\text{ROS}_{st}(R)$. The information on table is accumulated along the evaluation:
\begin{equation}
    \begin{aligned}
        \mathbf{D}_G(\sum^{n}\{R^{(k)}_n\}_{k=1}^K) \triangleq \text{SET}_{n-1} \oplus \sum_{k}^K \text{ROS}_{st}(R^{(k)}_n) 
    \end{aligned}
    \label{eqn_DG}
\end{equation}
Since a failed run $R$ implies that the corresponding $\text{ROS}_{st}(R)$ has not efficiently explored, it is not accumulated on $\text{SET}$. 

\begin{figure}[!t]
    \centering
    \includegraphics[width=\linewidth]{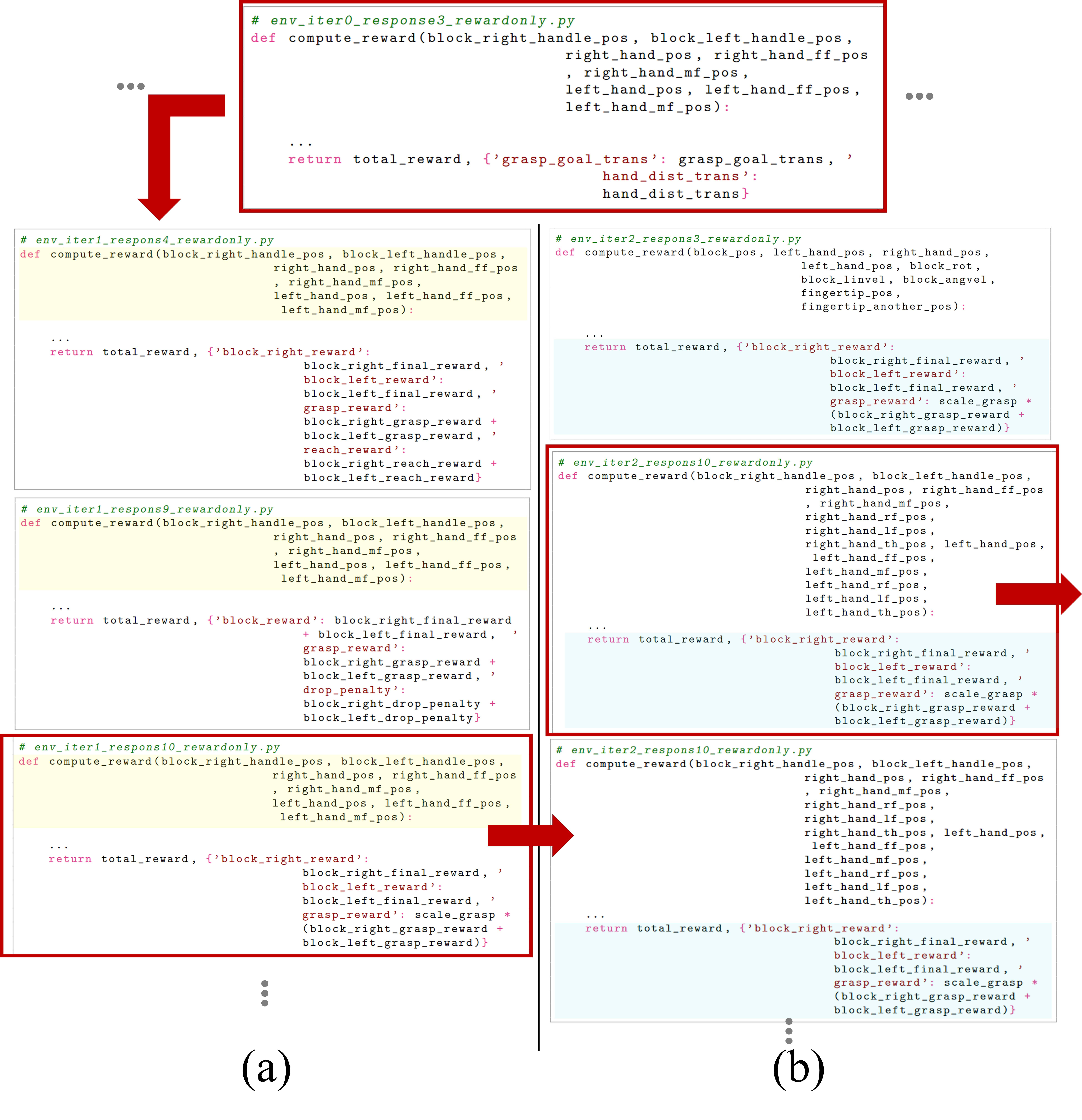}
    \vspace{-8mm}
    \caption{\textbf{Reward evolution in the first 3 iterations of our framework on \texttt{BlockGrasp} task.} (a) Each sample in iteration 2 keeps the same $\text{ROS}_{st}$ as $R_1^\star$. (b) Each sample in iteration 3 keeps the similar $\text{ROS}_{op}$ as $R_2^\star$. In each iteration, the abstract part the highest $\mathcal{F}_{sc}(\cdot)$ (identified by the red box) and two additional executable rewards are shown. }
    \vspace{-4mm}
    \label{fig_evolution}
\end{figure}

\subsection{Implementation details}

\noindent\textbf{LLM settings.} 
We use GPT-4, in particular the \texttt{gpt-4-0314} variant, as the backbone API for both $\text{LLM}_{R}$ and $\text{LLM}_{C}$. Besides passing $\mathbf{D}_U(U) $, the two LLMs do not share context with each other. During prompting, the temperature is set as $1.0$ to keep the diversity of reward samples. Each evolution is set to $N=5$ iterations, and LLM is required to generate $K=16$ samples simultaneously in each $\text{LLM}_{R}$ design iteration. $\text{ROS}_{st}$ is implemented by retaining only the first and last logical line of the reward code. $\tau$ reflects the difficulty of the given task. ~~The success threshold $\tau$ for each task is pre-set as the average success rate of $K_0=16$ independent and executable rewards designed by $\text{LLM}_{R}$. All prompts and task-independent templates are are attached to our \supmat. 

\noindent\textbf{RL settings.}
All environment code $M$ are deployed as IsaacGym~\cite{makoviychuk2021isaac} environments, which can be simulated simultaneously with high efficiency. Proximal Policy Optimization (PPO) algorithm~\cite{schulman2017proximal}, which uses default training parameters in the environments~\cite{chen2022towards}, is used as a practice for RL algorithms $\varLambda$. The rewards designed in the same iteration are distributed and trained on 8 NVIDIA RTX 3090 GPUs. The optimal reward candidate $R_N^\star$ are evaluated in 5 independent environments. The maximum interaction epochs per $\varLambda$ is set to 3000 in each $\text{LLM}_{R}$ design process, and 6000 in each evaluating process. 
\begin{table*}[!t]
    \caption{\textbf{Ablation studies on 4 Bi-dexterous Manipulation tasks.} From top to bottom, each row reports a variant framework that makes only one modification \wrt our final version (the last row) as specified by its name. }
    \setlength\tabcolsep{4pt}
    \rowcolors{1}{}{lightgray}
    \vspace{-4mm}
    \label{tab_eureka_ablation}
    \begin{center}
        \resizebox{0.8\linewidth}{!}{
            \begin{tabular}{l|l|l:l|l:l|l:l|l:l}
            \noalign{\hrule height 1.5pt} 
            && \multicolumn{2}{c}{\texttt{BlockStack}} 
            & \multicolumn{2}{c}{\texttt{DoorCloseOutward}} 
            & \multicolumn{2}{c}{\texttt{CatchAbreast}} 
            & \multicolumn{2}{c}{\texttt{Pen}}  \\
            \cmidrule{3-10}
            Index & Variants& \emph{$ESR_{avg}$}$\uparrow$ &\emph{$SSD$}$\uparrow$ &
            \emph{$ESR_{avg}$}$\uparrow$ &\emph{$SSD$}$\uparrow$ &
            \emph{$ESR_{avg}$}$\uparrow$ &\emph{$SSD$}$\uparrow$ &
            \emph{$ESR_{avg}$}$\uparrow$ &\emph{$SSD$}$\uparrow$ \\
            \midrule
            1 & Baseline$+\mathbf{D}_U(\cdot)$ & 0.09 & 0.37
                        & 0.37 & 0.39
                        & 0.00 & 0.30
                        & 0.11 & 0.24\\
            2 & 1$+\text{SET}$ & 0.12 & 0.39
                        & 0.59 & 0.74
                        & 0.00 & 0.53
                        & 0.05 & 0.74\\
            3 & 2$+\mathbf{D}_R(\cdot)^{-}$ & 0.13 & \textbf{0.48}
                        & \textbf{1.00} & 0.74
                        & \textbf{0.54} & 0.53
                        & 0.70 & 0.74\\
            4 & 2$+\mathbf{D}_R(\cdot)$ & \textbf{0.35} & 0.41
                        & 0.42 & \textbf{0.81}
                        & 0.00 & 0.53
                        & \textbf{0.80} & 0.57\\
            \noalign{\hrule height 1.5pt} 
            \end{tabular}
        }
    \end{center}
    \vspace{-4mm}
\end{table*}

\begin{figure*}[!t]
    \centering
    \includegraphics[width=0.8\linewidth]{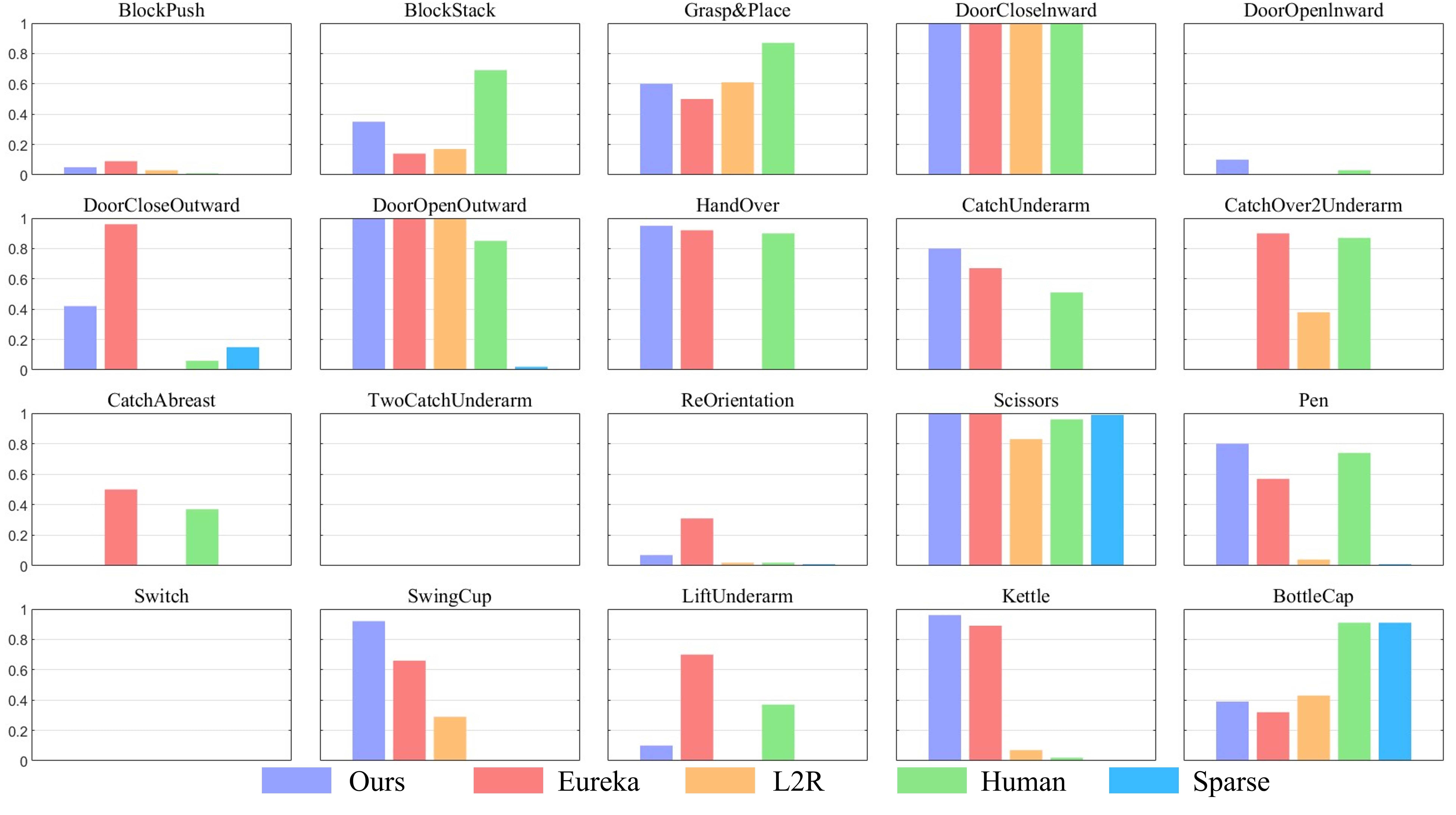}
    \vspace{-4mm}
    \caption{\textbf{Comparison with existing LLM reward design approaches.} The subgraphs report the success rates on the 20 dexterity tasks on the Bi-dexterous Manipulation benchmark~\cite{chen2022towards}.}
    \vspace{-4mm}
    \label{fig_compare}
\end{figure*}

\vspace{-0.2cm}
\section{Experiment}
\vspace{-0.1cm}
\subsection{Baselines and Tasks}

\noindent\textbf{L2R}~\cite{yu2023language} is a non-iterative framework that relies on reward function templates. For environments and tasks specified in natural language, the first LLM was asked to fill in a natural language template describing the agent's movements; the second LLM was then asked to convert this ``action description'' into code that invoked a set of manually-defined reward API primitives to write a parameterized reward program. To make the L2R competitive to our work, we follow ~\cite{ma2023eureka} to define motion description templates to mimic the original L2R templates, Its API reward primitives were constructed using individual components of the original human rewards.

\noindent\textbf{Eureka}~\cite{ma2023eureka} utilizes the code-writing and zero-sample-generation capabilities of Large Language Models (LLMs) to generate executable reward function code directly from the environment source code and linguistic task descriptions in an iterative framework. This process does not require task-specific hints, enabling Eureka to achieve zero-sample reward function generation across a wide range of tasks. Eureka employs an evolutionary search strategy by iteratively sampling reward functions, evaluating their performance, and incrementally improving the reward functions based on feedback from these evaluations. For the fairness of the comparison, our approach is consistent with Eureka in terms of iteration amount $N=5$ and sample amount $K=16$. 

The above baselines and our framework are evaluated on 20 hand-object interaction task from Bi-dexterous Manipulation benchmark~\cite{chen2022towards} with the same hyperparameter settings. These task can be further categorized as: (i) Block Manipulation ($\times3$): BlockPush, BlockStack, Grasp\&Place. (ii) Door Manipulation ($\times4$): DoorCloseOutward, DoorCloseInward, DoorOpenOutward, DoorOpenInward. (iii) In-hand Manipulation ($\times6$): ReOrientation, HandOver, CatchUnderarm, CatchOver2Underarm, CatchAbreast, TwoCatchUnderarm. (iv) Functional Grasp ($\times8$): Pen, Switch, GraspAndPlace, Kettle, Scissors, SwingCup, BottleCap, LiftUnderarm. It is noted that there are no templates or modules specifically designed for the Bi-dexterous Manipulation in our framework. The 9 tasks in IsaacGymEnvs~\cite{makoviychuk2021isaac} are ignored in our experiments, mainly stemming from the fact that they contain relatively few environment states and that Eureka can already design better forms of rewards.

\subsection{Evaluation Metrics}

The following metrics are used to quantitatively evaluate the quality of the different reward design approaches in each task. It is worth noting that the quantities between different tasks are relatively independent and not comparable. 

\noindent\textbf{Evaluated success rate} (\emph{$ESR$}) for each task indicates the maximum success achieved in once RL training procedure with a given reward. Consistent with Eureka~\cite{ma2023eureka}, 5 identical but independent RL training procedures are launched after each reward design procedure. The average of the 5 rates (\emph{$ESR_{avg}$}) is adopted to measure the effective of the given reward and a higher \emph{$ESR_{avg}$} means a more effective reward.

\noindent\textbf{Sampling state disparity} (\emph{$SSD$}) for each task refers to the bias between the maximum and average usage frequency of all states for historical reward codes (whether successfully executed or not). It reflects the exploration degree of a reward design approach on all states within the task. 

\subsection{Comparisons}

\figmk\ref{fig_compare} reports the performance comparison of our framework with other existing approaches on 20 tasks. \emph{$ESR_{avg}$} is used as the evaluation metric. According to the results in the table, our method performs better than Eureka in 9 tasks and matches Eureka's performance in 5 extreme simple/difficult tasks. Only in 6 tasks (BlockPush, CatchAbreast, CatchOver2Underarm, DoorCloseOutward, ReOrientation, LiftUnderarm), our method performs slightly lower than Eureka. 

We further found that it was the success rate thresholds for these tasks that were set too high, causing $\text{LLM}_R$ to focus too much on space member selection throughout the evolution process. In our ablation study, two underperforming tasks (CatchAbreast and DoorCloseOutward) as well as two outperforming tasks are selected. However, in 2$+\mathbf{D}_R(\cdot)^{-}$ where the threshold is set to 0.1, we outperformed Eureka on both underperforming tasks, achieving scores of [0.54, 1], which are higher than Eureka's scores of [0.5, 0.96].  This suggests that adjusting thresholds might have a significant impact on improving learning efficiency and task performance across various robotic tasks. In our current approach, we use the average of the success rates achieved by LLM in the first iteration of the reward design as the average success rate for each task. In the future, an adaptive threshold search algorithm could enable this mechanism to achieve better performance.

\vspace{-0.1cm}
\subsection{Ablation Study}
Starting with the Baseline$+\mathbf{D}_U(\cdot)$, we observe that the naive introduction of the user-expert mission reconciling mechanism can marginally improve LLM reward design. This means that only formal and informative requirements descriptions are not sufficient for LLM based reward evolution. When the state execution table (SET) is introduced (1$+\text{SET}$), we observe a more consistent improvement, especially in the \texttt{DoorCloseOutward} task where \emph{$ESR_{avg}$} increases from 0.37 to 0.59, and \emph{$SSD$} improves significantly. This suggests that incorporating structured state execution improves the system’s ability to handle more complex scenarios by providing a clearer execution plan. The addition of a fixed success threshold $\tau=0.1$ in variant (iii), 2$+\mathbf{D}_R(\cdot)^{-}$, further boosts performance, particularly in tasks like \texttt{BlockStack} and \texttt{CatchAbreast}, where \emph{$ESR_{avg}$} and \emph{$SSD$} either match or exceed prior results. The fixed threshold likely provides a clearer metric for success, leading to more stable training and execution. Finally, variant (iv), 2$+\mathbf{D}_R(\cdot)$, which introduces a task-specific success threshold via automatic hyperparameter search, results in significant improvements in tasks like \texttt{BlockStack} (with an \emph{$ESR_{avg}$} of 0.35) and \texttt{Pen} (with an \emph{$ESR_{avg}$} of 0.80). However, the improvement is not uniform across all tasks, as the \texttt{CatchAbreast} task continues to show little progress, indicating that task-specific thresholds may not universally enhance performance across all environments.

\vspace{-0.2cm}
\section{Conclusion}
\vspace{-0.1cm}
In this work, we introduced a novel framework that enhances the design of reinforcement learning (RL) rewards by harnessing the capabilities of large language models (LLMs). Our approach addresses the critical challenge of grounding, facilitating more effective communication between LLMs and RL agents by evolving the Reward Observation Space (\rwdobs) through heuristic sampling. By incorporating a table-based exploration caching mechanism, we alleviate the Markovian constraint commonly found in LLM dialogues, enabling a more comprehensive and efficient exploration of potential reward spaces. Additionally, our structured text-code reconciliation strategy bridges the cognitive gap between user intentions and expert-defined success criteria, ensuring that LLM-generated rewards are both relevant and actionable across a variety of environments. The proposed method surpasses existing frameworks in both efficiency and effectiveness, highlighting its potential for universal RL reward design. This research represents a notable advancement in integrating LLMs with RL, paving the way for the development of autonomous systems that can adapt to diverse tasks with minimal human input. Future work will aim to further refine the feedback loop and investigate the application of this framework in more complex, real-world scenarios.

\bibliographystyle{plain}
\bibliography{reference}

\end{document}